\documentclass[ms]{informs3} 

\OneAndAHalfSpacedXII 

\usepackage{natbib}
 \bibpunct[, ]{(}{)}{,}{a}{}{,}%
 \def\bibfont{\small}%
\TheoremsNumberedThrough     

\EquationsNumberedThrough    

\usepackage[toc,title,page]{appendix}
\usepackage{booktabs}
\usepackage{amsmath}
\usepackage{amssymb}
\usepackage{babel}
\usepackage{bbold}
\usepackage{comment}
\usepackage{graphics}
\usepackage{graphicx}
\usepackage{placeins}
\usepackage{tikz}
\usetikzlibrary{automata,arrows,positioning,calc}
\usepackage{color}
\usepackage{algorithm}
\usepackage{bbold}
\usepackage{paralist}
\usepackage{mathtools}
\usepackage{xcolor}
\usepackage{multirow}
\usepackage{algcompatible}
\usepackage{caption}
\usepackage{subcaption}
\usepackage{multirow}

\newcommand{\gdim}{\mathrm{Graph}}
\newcommand{\ndim}{\mathrm{Natarajan}}
\newcommand{\vcdim}{\mathrm{VC}}

\newcommand{\covering}{N_H}
\newcommand{\packing}{M_H}
\newcommand{\ei}{\kappa} 

\newcommand{\ipw}{\hat{\pi}_{\text{AIPW}}}

\newcommand{\feas}{\mathcal{X}}
\newcommand{\actions}{\mathcal{A}}

\newcommand{\W}{A}
\newcommand{\QE}{\hat{Q}}

\newcounter{proofstep}

\begin{document}


\RUNAUTHOR{Qu, Qian, and Zhou}

\RUNTITLE{Interpretable Policy Learning}

\TITLE{Interpretable Personalization via Policy Learning \\with Linear Decision Boundaries}


\ARTICLEAUTHORS{%
\AUTHOR{Zhaonan Qu}
\AFF{Department of Economics, Stanford University, \EMAIL{zhaonanq@stanford.edu},
\URL{}}
\AUTHOR{Isabella Qian}
\AFF{University of California, Los Angeles, \EMAIL{iqian@ucla.edu}, 
\URL{}}
\AUTHOR{Zhengyuan Zhou}
\AFF{Stern School of Business, New York University, \EMAIL{zzhou@stern.nyu.edu}, \URL{}}
} 

\ABSTRACT{%
\textbf{Problem definition:} With the rise of the digital economy and an explosion of available information about consumers, effective personalization of goods and services has become a core business focus for companies to improve revenues and maintain a competitive edge. This paper studies the personalization problem through the lens of policy learning, where the goal is to learn a decision-making rule (a policy) that maps from consumer and product characteristics (features) to recommendations/treatments (actions) in order to optimize outcomes (rewards). 
We focus on using available historical data for offline learning with unknown data collection procedures, where a key challenge is the non-random assignment of recommendations. Moreover, in many business and medical applications, \emph{interpretability} of a policy is essential. \textbf{Methodology/results:} We study the class of policies with linear decision boundaries to ensure interpretability, and propose learning algorithms using tools from causal inference to address unbalanced treatments. We propose several optimization schemes to solve the associated non-convex, non-smooth optimization problem, and find that a Bayesian optimization algorithm is fast and effective. We test our algorithm with extensive simulation studies and apply it to an anonymized online marketplace customer purchase dataset, where the learned policy outputs a personalized discount recommendation based on customer and product features in order to maximize gross merchandise value (GMV) for sellers. Our learned policy improves upon the platform's baseline by 88.2\% in net sales revenue, while also providing informative insights on which features are important for the decision-making process, e.g., when ``Attribute 2'' is large, marginal increase in GMV is low for discounts higher than 10\%. \textbf{Managerial implications:} Our findings suggest that our proposed policy learning framework using tools from causal inference and Bayesian optimization provides a promising practical approach to interpretable personalization across a wide range of applications.
}%


\maketitle

%

\vspace{-1cm}
\section{Introduction}
\label{sec:intro}

An increasingly prominent feature of the digital economy is the personalization
of goods and services based on consumer, product, and contextual information to improve customer experience and business revenues \citep{iyer2005targeting,li2010contextual,kim2011battle,kallus2021minimax,kallus2021fairness,bastani2022meta,bastani2021learning,decroix2021service}. In online platforms, product recommendations, advertisements, and offers are often based on customers' characteristics, preferences, and browsing histories \citep{ettl2020data,pauphilet2022robust}. Similarly, in precision medicine and pharmacogenetics \citep{bertsimas2017personalized,hespanhol2018family, zhou2018evaluating, drew2016pharmacogenetics,minor2020discovering,bastani2020online}, different treatment options and dosages can be selected
based on patients' medical histories and genetic profiles in order to optimize patients' outcomes. Even in the more traditional domain of public policy,
individual heterogeneities are becoming increasingly important when
determining what kind of policies should be applied to particular groups in order to balance optimizing individual outcomes with improving overall social welfare or reducing inequality \citep{kitagawa2017equality}. 

The key phenomenon underlying the ubiquity of personalization across different domains is \emph{heterogeneity}, that is, different individuals respond differently to particular recommendations/offers/treatments. As a result, the best option may vary significantly across individuals in a population. To quote a proverb provocatively: ``One man's poison is another man's medicine''. Thanks to the growing availability of historical data on consumers, users, and patients, it is now possible to make better decisions based on such individual heterogeneities, and personalization has transformed from heuristics-based rules to sophisticated data-driven operations using increasingly detailed information.

To enable good personalization from \emph{historical} data, the central problem we need to solve is that of \emph{offline} policy learning: to
learn a ``good'' personalized decision-making rule (a policy) that maps individual and contextual information to available actions, in order to achieve better outcomes, such as
higher click-through rates for ads, increased sales and revenue in online marketplaces, and improved
patient outcomes. 
One of our motivating applications comes from the e-commerce platform JD.com \citep{shen2020jd}, which released large amounts of anonymized historical data on customer purchase behavior that contain information on customer and product characteristics (features/contexts),  discounts available to customers at time of purchase (treatments/actions), and customers' purchase decisions and sales value (rewards). A natural question is whether the platform and third-party sellers can improve net sales revenue by targeting and personalizing discount offers to potential buyers in order to encourage them to make a purchase decision. Because of variations in the availability and amount of discounts, 
the historical dataset may already contain enough heterogeneity that allows sellers to learn a good policy \emph{offline}, which recommends an optimal discount offer to each potential customer considering a particular product, without the need to conduct further online experiments that could take time and result in revenue losses. Such a policy learned using historical data provides a good operational starting point for personalized discount offers, and could also help the platform/sellers decide whether further online experiments are worthwhile. See also \cite{ettl2020data} for a related but different problem on personalizing discounted product \emph{bundles} based on online consumer preferences and inventory constraints.


Despite the appeal of offline policy learning, the \emph{observational} nature of the data also poses unique challenges. Most importantly, unlike in online policy learning, we do not have control over treatment assignments. Outcomes observed in historical data are results of the particular treatments selected, and the exact treatment assignment mechanism(s) used to collect the data may not be available information to us. In the purchase dataset from JD.com, some discount offers were assigned randomly by the platform, while others had to be claimed (``clipped'') first by customers before applying them. Yet other discounts were already results of targeting based on customer information using the platform's machine learning algorithms \citep{shen2020jd}. To learn from these diverse sources of data derived from different assignment mechanisms, a learning algorithm needs to handle selection bias (a.k.a. confounding) on the treatments observed, i.e., the assignment of discount offers can be correlated with customers' potential purchase outcomes. This is certainly the case for targeted offers, but also for discounts actively claimed by customers, whose characteristics are correlated with discount-seeking behavior as well as purchase decisions. In the presence of confounding, simple supervised learning approaches using the observational dataset will result in biased estimates for rewards. Similarly, in a hospital setting \citep{bennett2020efficient}, since healthier patients (who are more likely to recover) tend to receive less invasive treatments, failure to address selection bias leads one to conclude that less invasive treatments always result in better patient outcomes.



Fortunately, the challenges posed by the observational nature of historical data are especially amenable to tools from causal inference \citep{imbens2015causal}, and in particular the idea of \emph{propensity weighting} \citep{horvitz1952generalization,rosenbaum1983central,robins1994estimation,hirano2003efficient}, whereby samples that are assigned to less common treatment arms are given more weight when forming sample average estimates of treatment effects. Building on semiparametric causal estimates of potential rewards, a series of recent works on offline policy learning focus on optimal regret bound and sample complexity of general policy classes, e.g., \cite{athey2017efficient,kitagawa2015should,zhou2018offline,demirer2019semi,kallus2021minimax}. However, despite the success of efficient causal estimates in providing theoretical guarantees for offline policy learning algorithms, practical considerations for operationally relevant policy classes remain to be explored further. In particular, many business and medical applications demand that learned policies be easy to understand and \emph{interpret}, so that practitioners can assess and make sense of the relative importance and contributions of each feature in critical decision-making processes.

In this paper, we address the need for practical and interpretable personalization via offline policy learning and propose an end-to-end learning framework for policies with linear decision boundaries, where each treatment arm is parameterized by a linear model and the policy assigns an individual to the treatment arm with the largest model value given the individual's features. Instead of approximating \emph{rewards} directly with linear models in order to achieve interpretability, as is commonly done in the literature, our framework allows the use of flexible blackbox (and non-linear) reward estimators, and parameterizes the \emph{policies} by linear functions in order to achieve interpretability. This decouples reward estimates from policy functions, and allows us to incorporate semiparametric estimates of rewards from causal inference in order to address challenges of selection bias in observational datasets as well as nonlinear dependence of rewards on features. Linear decision boundaries are arguably the most natural and simple policy class to use in practice, as they allow us to assess the relative importance of each feature in making a decision. 
Despite this simplicity, the optimization problem associated with the offline learning problem poses significant challenges due to the non-smooth and non-convex nature of the objective, which we will address in the present work using Bayesian optimization.

\subsection{Our Message} 
Our philosophy is ``Complex Environment, Interpretable Policy", which entails first obtaining sophisticated 
estimates of the environment (using state-of-the-art blackbox estimation schemes) and then optimizing for the best interpretable policy using these estimates. This is different from--and superior to--making interpretable (e.g. linear) model assumptions on the environment and then solving for the best policy based on that (``Simple Environment, Simple Policy''), which is the popular choice in applications that prioritize interpretability. Our approach also differs from decision rules directly based on complex models of the environment (``Complex Environment, Complex Policy'') in that it offers interpretability, for which the practical significance is well understood (see for instance the excellent recent work of~\cite{bastani2022interpretable}) and enjoys theoretical guarantees on regret.

Our main takeaways are twofold.
First, when finding policies with linear decision boundaries, the interpretable policy class we focus on in this paper, it is crucial to solve the empirical reward maximization (ERM) problem directly rather than through any relaxation. This is because the ERM problem is built using highly accurate (blackbox) reward estimates, which lose their power if the ERM problem is not solved exactly. In our setting, this ERM problem is non-convex and non-smooth, hence posing a computational challenge if solved in a brute-force way. By adapting a general Bayesian optimization methodology~\citep{zhou2014gradient} to our problem, we construct an algorithm (Algorithm~\ref{alg:GASS}) that provides an attractive computational option while retaining statistical efficiency. In comparison to the folklore convex relaxation and logistic relaxation approaches that may seem to be the most obvious candidates 
to address the challenges of non-convex and non-smooth problems, our Bayesian algorithm performs much better empirically (see simulations in Section~\ref{sec:optimization}) and is robust to noise. More broadly, our results suggest that Bayesian Optimization, which currently appears to be less popular in the policy learning community, can be an effective tool in the toolkit.

Second, to demonstrate the practical appeal of our learning framework as a decision-making tool for interpretable personalization to a broad range of problems, we apply our algorithm to a customer purchase dataset provided by JD.com \citep{shen2020jd}, China's largest online retailer. Our algorithm learns a policy that recommends personalized discount offers to consumers based on consumer, product, and other contextual information to maximize sellers' gross merchandise value (GMV). A simple implementation using a Bayesian optimization approach suggests an increase in net sales revenue by 88.2\% over the current baseline used by sellers. The learned policy also provides interpretatable insights into how customer and product characteristics determine the impact of discounts, which helps inform the platform on which features are important for improving GMV.  For example, we found that the marginal return in net sales revenue from discounts is relatively small when the quality indicator ``Attribute 2'' of a product is high, with optimal discount above 0\% but below 10\%, especially for users with a ``plus'' membership. 

\subsection{Related Works}

\textbf{Policy Learning.} The problem of policy learning from observational data, also known as optimal treatment allocation, has been studied simultaneously in several different disciplines, such as economics \citep{manski2004statistical,manski2009identification,hirano2009asymptotics,kitagawa2015should,athey2017efficient}, machine learning and operations research \citep{dudik2011doubly,swaminathan2015batch,zhou2018offline,kallus2021minimax}, and statistics \citep{zhang2012estimating,luedtke2016statistical,zhao2012estimating}. From a theoretical perspetive, the central measure of the quality of a policy is the asymptotic regret, i.e., the difference in expected outcomes associated with the optimal decision rule and that provided by the policy, as sample size grows large. To achieve the minimax optimal  $\mathcal{O}_p(1/\sqrt{n})$ asymptotic regret when the data collection mechanism is unknown and endogenous, most works with such guarantees, such as \cite{athey2017efficient,kallus2021minimax,zhou2018offline}, rely crucially on the doubly robust estimators of \cite{robins1994estimation} and the associated statistical theory on semiparametric estimation \citep{hahn1998role,newey1994asymptotic,robins1995semiparametric,imbens2004nonparametric}, although doubly robust estimators have already been used (without guarantees) for policy learning in applications such as healthcare \citep{tsiatis2011improved}, education \citep{uysal2015doubly}, and online recommendations \citep{wang2019doubly}.
Recent works have also examined offline policy learning when the data distribution is not stationary. One idea is that \emph{adaptive} data collection could improve offline policy learning \citep{zhan2021policy,kallus2021more,bibaut2021risk}. Other works consider individuals reacting strategically to treatments \citep{munro2020learning} or competing against each other under capacity constraints \citep{sahoo2022policy}. Although it would be interesting to examine the extent to which consumers react strategically to personalized treatments, in this paper, we will assume that the data distribution is stationary, and leave extensions to future work. 

\textbf{Bayesian Optimization.}
Although itself an established research field, the use of Bayesian optimization has been less common in policy learning. Our paper builds upon the line of works aiming to combine model-based Bayesian algorithms, which consider a population of potential solutions, with gradient-based algorithms, which enjoy fast convergence. In particular, \cite{zhou2014gradient} propose a general framework based on a stochastic quasi-Newton iteration, which iterates between sampling from the domain of solutions and updating the \emph{distribution} on the solution space. The distribution eventually concentrates on the set of global optima for the original problem. A major advantage of this approach is its ability to overcome local optima, which methods based on updates of a single solution often suffer from. Moreover, it is applicable to non-smooth objectives, as the original objective function is replaced with a smooth problem over the parameter space of the sampling distribution. Importantly, these algorithms can often be viewed as a generalized Robbins-Monro algorithm \citep{robbins1951stochastic,chung1954stochastic}, whose convergence can be studied with tools from stochastic approximation theory \citep{borkar2009stochastic}.
\section{Problem Setup} \label{sec:model}

In this section, we introduce the offline policy learning framework used to learn decision rules for personalization from observational data in this paper.
Let $\mathcal{X} \subset \mathbf{R}^p$ be the underlying context space of dimension $p$ and
$\mathcal{A}=\{a^{1},\dots,a^{d}\}$ be a set of $d(\geq2)$ actions (or treatments).
We posit that there is an underlying distribution\emph{
$\mathbb{P}$} on $\big(X,Y(a^{1}),Y(a^{2}),\dots,Y(a^{d})\big)\in\mathcal{X}\times\prod_{i=1}^{d}\mathcal{Y}_{i}$, where $Y(a^j) \in \mathcal{Y}_j \subset \mathbf{R}$ denotes the 
\textbf{potential outcome} that represents the random reward \emph{if} the corresponding action $a^j$ \emph{were} taken~\citep{rubin1974estimating, imbens2015causal}.
We assume that all $\mathcal{Y}_j$'s are bounded.
Denote the expected reward as $\mu(x,a):=\mathbb{E}[Y(a)\mid X=x]$, which can have arbitrary non-linear dependence on $x$. In the example of targeted discount offers in online markets, $\mu(x,a)$ could represent the expected net sales revenue for the platform if a discount of type $a$ were recommended to a customer-product pair $x$.
 
We are given a dataset consisting of $n$ i.i.d. triples $\{(X_{i},A_{i},Y_{i})\}_{i=1}^{n}$ sampled as follows. First, $(X_i,Y_i(a^{1}),Y_i(a^{2}),\dots,Y_i(a^{d}))$ is sampled i.i.d. from $\mathbb{P}$; and then conditional on the context $X_i$, action $A_i\in \mathcal{A}$ is sampled using a probability vector of \textbf{propensities} $\big(e(X_i, a^1), \dots, e(X_i, a^d)\big)$ where $e(x, a^j) = \mathbb{P}[A_i=a^j\mid X=x]$, after which \emph{only} $Y_i=Y_i(A_i)$ is revealed. Importantly, the data-collection rule underlying the propensity model is generally unknown to us, and could be a simple heuristic or a sophisticated machine learning model (or a combination of both). However, we make the following standard assumption in the causal inference literature that it only depends on observable $x$ and not on any $Y(a^j)$.

\begin{assumption}[Unconfoundedness]
\label{assump:unconfounded}
$A_i \perp (Y_i(a^{1}),Y_i(a^{2}),\dots,Y_i(a^{d})) \mid X_i$
\end{assumption}

We emphasize that we
only observe the reward associated with the \emph{realized} action $A_{i}$ actually selected in the  $i^{\mbox{th}}$ observation, even though all the other unobserved potential
outcomes $Y_{i}(a)$ for  $a\in\mathcal{A}\setminus\{A_{i}\}$---also known as \textbf{counterfactuals} \citep{imbens2015causal}---exist in the
model and have been drawn in accordance with  $\mathbb{P}$. The issue of missing counterfactuals poses great challenges for observational studies  and is mitigated with the ``overlap'' assumption \citep{robins1995semiparametric,hirano2003efficient,chernozhukov2018double} that the data-collection rule has sufficient exploration, which is essential for accurately estimating the reward functions and learning policies.
\begin{assumption}[Overlap]
\label{assump:classical}
There exists some $\eta>0$, such that the propensity $e(x,a)\geq\eta$, for any $(x,a)\in\mathcal{X}\times\mathcal{A}$. 
\end{assumption}

The problem of \textbf{offline policy learning} is to learn a \emph{policy} $\pi$, which maps a context vector $x\in\mathcal{X}$ to an action choice $a\in \mathcal{A}$, from the observational data $\{(X_{i},A_{i},Y_{i})\}_{i=1}^{n}$ in order to maximize reward. Specifically, given a policy $\pi$,
its expected \textbf{policy value} is
\begin{align}
Q(\pi) & :=\mathbb{E}[Y_{i}(\pi(X_{i}))], 
\label{eq:expected reward}
\end{align}
where the expectation is taken over the data generating process.
The \textbf{optimal policy} in a function class $\Pi$ is defined
as $\pi^{\ast}:=\arg\max_{\pi\in\Pi}Q(\pi)$.
We measure the performance of a policy $\pi \in \Pi$ by its \textbf{regret}, which is its value gap from the optimal policy:
\begin{align}
R_{\Pi}(\pi) & :=Q(\pi^{\ast})-Q(\pi)\label{eq:regret} =\mathbb{E}\left[Y(\pi^{\ast}(X))\right]-\mathbb{E}\left[Y(\pi(X))\right].
\end{align}
Our goal is to learn a policy $\hat{\pi}$ from $\Pi$ such that the regret of $\hat{\pi}$ is small \emph{with high probability}. Here the randomness comes the fact that $\hat{\pi}$  is a function of the stochastically collected observational dataset. Note, however, that in this paper, we consider the underlying distribution $\mathbb{P}$ to be stationary and the reward to be non-adversarial, so it suffices to learn a deterministic policy that maps each context to an action choice, as opposed to a random policy whose range is the probability simplex over the action set $\mathcal{A}$.  

\textbf{Remark.}  We make two comments that help distinguish offline policy learning from related problems and approaches. First, although both online and offline policy learning use regret as their performance metric, they are fundamentally different problems. In online policy learning problems such as contextual bandits \citep{agrawal2013thompson,bastani2020online}, the key challenge is balancing the exploration-exploitation trade-off by carefully designing the treatment assignment mechanism. In offline policy learning problems, the exact treatment assignment mechanism used in data collection was pre-determined and may \emph{not} be available to us, but we assume that it only uses observable contexts and affords enough exploration to learn a good policy. Second, one may consider directly estimating $\mu(x,a)$ for each $a$ with regression or machine learning methods, using observations $(X_{i},A_{i},Y_{i})$ where $A_i=a$. However, as long as $A_i$ depends on $x$, this will result in biased estimates due to confounding. In the JD.com example, the coupon offer $a$ could have been targeted at customers who are deemed more likely to spend more money. Naive regression approaches would bias $\mu(x,a)$ upwards. Therefore, offline policy learning cannot be solved simply with traditional supervised learning methods.

\subsection{Learning Linear Decision Boundaries}
\label{sec:linear class}

Having set the stage for general offline policy learning, we now focus on a particular policy class $\Pi_\text{linear}$, known as \textbf{linear decision boundaries}, which are policies of the form:
\begin{align}
\pi(x; \Theta) & \in\arg\max_{a\in\mathcal{A}}\langle\theta_{a},x\rangle, \label{eq:linear policy class}
\end{align}
where $\theta_a\in \mathbb{R}^p$ for each $a\in\mathcal{A}$, i.e., each action is associated with a parameter vector $\theta_{a}$, and $\Theta \in \mathbb{R}^{pd}$ is obtained by stacking up $\{\theta_{a}\}_{a\in\mathcal{A}}$.

A main advantage of linear decision boundaries is that the coefficients can be interpreted as the relative importance of a given feature, which the learned policy places when trying to decide whether to employ a particular action or assign the unit to a particular treatment. For example, if the $j$-th component of $\theta_{a}$ is positive and large for a particular action $a$ while small in comparison or negative for other actions, then an increase in the $j$-th feature $x_j$ increases the likelihood of action $a$ being chosen, relative to other actions. In cases of decision-making  where interpretability is critical, such as precision medicine, this property is desirable, as doctors are able to deduce that particular patient characteristics positively or negatively influence the decision to adopt or avoid particular treatments. Similarly, in an online marketplace such as JD.com, a learned policy with linear decision boundaries for personalized discount recommendation provides insights to the platform and vendors on which consumer and product characteristics are important when determining how a particular amount or type of discount offer influences consumers' purchase behaviors. 

\begin{figure}
\begin{centering}
\includegraphics[scale=0.5]{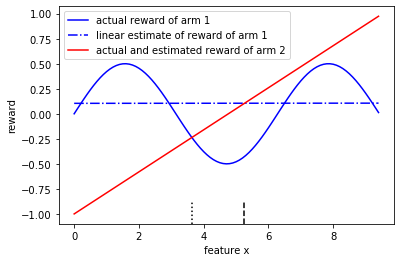}
\par\end{centering}
\caption{In the above example with two actions and one feature, arm 2 has a linear reward function, while arm 1 has a nonlinear reward function, whose linear estimate is the horizontal dot-dash line. The optimal linear decision boundary is the point on the x axis with the dotted vertical line above it. For feature values to the left of this point, arm 1 is the optimal assignment, while to the right arm 2 is optimal. Because the reward of arm 1 is nonlinear, any algorithm that assumes a linear structure for the reward function will instead estimate the horizontal approximation, based on which the decision boundary is the point on the x axis with the dashed vertical line above it. This decision boundary is strictly worse than the optimal linear decision boundary to its left. See Section \ref{sec:linear} for more examples.}
\label{fig:1dexample}
\end{figure}

\textbf{Remark.} Importantly, the policy class of linear decision boundaries  includes--as a \emph{strict} subset--policies based on linear models of rewards: whereas policies based on learning linear models of $\mu(x,a)$ place the linear structure assumption on rewards (which could easily suffer from model mis-specification, in addition to the confounding problem discussed before), $\Pi_\text{linear}$ only requires that decision boundaries to be linear, while allowing rewards to have arbitrary dependence on covariates. This is the crucial difference between our approach to interpretable personalization using linear decision boundaries and the approach of linear reward modeling, and embodies our philosophy of ``complex environment, simple and interpretable policy". In the special case when rewards indeed depend linearly on covariates, the optimal policy in $\Pi_\text{linear}$ coincides with the one based on linear models of rewards; in general, however, these two are different and the optimal policy in $\Pi_\text{linear}$ is strictly better. Figure \ref{fig:1dexample} provides a visualization of this difference in the toy example with two actions ($d=2$) and one-dimensional feature ($p=1$). We will illustrate this point further with simulations in settings where rewards depend non-linearly on features, but the optimal decisions are separated by linear boundaries in the feature space. 

\subsection{Policy Learning with Doubly Robust Reward Estimation}

For any policy $\pi$, given an observational dataset satisfying Assumption \ref{assump:classical}, we will approximate
its policy value $Q(\pi) =\mathbb{E}[Y_i(\pi(X_{i}))]$
  by the empirical estimator 
\begin{align}
    \widehat{Q}(\pi) = \frac{1}{n}\sum_{i=1}^n\widehat{\Gamma}_i( \pi(X_i)),
\end{align}
where $\widehat{\Gamma}_i(a)$ is an \emph{unbiased} estimator of the expected reward $\mathbb{E}[Y_i(a)|X_i]$.
The approach we take to learn an optimal policy in $\Pi_\text{linear}$ is by finding a solution to the following \textbf{empirical reward maximization} (ERM) problem:
 \begin{align}
\label{eq:ERM}
 \max_{\pi\in\Pi_\text{linear}}\hat{Q}(\pi)
 \iff \max_{\Theta\in\mathbb{R}^{dp}}\frac{1}{n}\sum_{i=1}^n\widehat{\Gamma}_i(\pi(X_i;\Theta)).
\end{align}

Here we raise two important questions of theoretical and practical interest, respectively. The first one is whether the solution to the ERM problem is \emph{optimal} in terms of asymptotic regret, while the second question concerns how to efficiently obtain a maximizer of the ERM problem.
The answer to the first question is affirmative and can be proved using tools from statistical learning theory, more specifically, regret bounds for multi-action policy classes via Natarajan dimensions \citep{jin2022upper}. In this paper, we will focus on the latter practical question of optimization, but will follow prior works on optimal regret bounds for policy learning, e.g., \cite{dudik2011doubly,athey2017efficient,kitagawa2017equality,zhou2018offline}, and consider the \emph{doubly robust} (a.k.a. augmented inverse propensity weighting or AIPW) estimators:
\begin{align}
\label{eq:AIPW}
\hat{\Gamma}_{i}(a) & =\frac{Y_{i}-\hat{\mu}(X_{i}, a)}{\hat{e}(X_{i}, a)}\cdot \mathbf{1}\{A_{i}=a\}+\hat{\mu}(X_{i},a)
\end{align}
where $\hat{\mu}$ and $\hat{e}$ are estimators
for expected rewards $\mu$ and propensities $e$.
 Here double robustness refers to the fact that the estimator is consistent if \emph{either} the estimators of propensities are consistent, \emph{or} the estimators of expected rewards are consistent. Moreover, doubly robust estimators are also semiparametrically efficient, i.e., they achieve the lowest asymptotic variance among all semiparametric estimators~\citep{robins1994estimation}. 
In principle, one can use different procedures to estimate the propensity
scores and expected rewards for different arms in $\mathcal{A}$, as long as the estimators satisfy certain rate conditions. In practice, even
though some non-linear estimators of $\mu$ and $e$, such as deep neural networks, may not
enjoy these rates, they may still be advantageous and yield
good empirical performance if the propensity or reward models
are highly nonlinear as functions of the features. We reiterate that the use of non-linear estimators for reward does not affect the interpretability of our learned policies, because the linearity assumption is placed on the policy's decision boundaries, not on the rewards.
%

Regarding the second question of how to obtain a maximizer of the ERM problem \eqref{eq:ERM}, the short answer is that it is in general computationally hard, as the optimization objective
is non-differentiable and non-convex due to the argmax component of policies with linear decision boundaries.
In Section \ref{sec:optimization}, we present several practical approaches to find good solutions. Before that, we will first prove the minimax optimality, in terms of asymptotic regret in $\Pi_{\text{linear}}$, of maximizers of the ERM problem \eqref{eq:ERM} with doubly robust reward estimators. Readers interested in applications can skip the next chapter.

\section{Bayesian Optimization for Learning Linear Decision Boundaries}
\label{sec:optimization}

We now move on to solving the ERM problem $\max_{\Theta\in\mathbb{R}^{dp}}\frac{1}{n}\sum_{i=1}^n\widehat{\Gamma}_i(\pi(X_i;\Theta))$ defined in \eqref{eq:ERM}, with the goal of learning the optimal policy with linear decision boundaries from the  observational data. It is challenging because the associated ERM problem 
is non-differentiable and non-convex due to the argmax components in linear decision boundaries $\pi(x;\Theta) \in\arg\max_{a\in\mathcal{A}}\langle\theta_{a},x\rangle$. 

To address these difficulties, prior works have focused on relaxation approaches~\citep{shalev2014understanding,bennett2020efficient}. In particular, the ERM problem can be interpreted as a \emph{weighted} classification problem, which can then be relaxed to a multiclass support vector machine. The convex relaxation approach constructs a concave lower-bound of the original problem and then maximizes it via standard techniques that find the global optima of the lower bound \citep{shalev2014understanding}. 
Alternatively, we also propose searching in the space of multinomial \emph{logistic} policy class $\Pi_{\text{logistic}}$,
where the random policies can be obtained via a softmax mapping:
\begin{align}
    \Pi_{\text{logistic}} = \big\{h(x;\Theta) = \mbox{Softmax}(x^T\theta_{a^1}, \dots, x^T\theta_{a^d})\big\}.
\end{align}
The solution in $\Pi_{\text{logistic}}$ is then projected into $\Pi_{\text{linear}}$ to obtain the final learned policy.
Such a logistic relaxation yields a differentiable but still non-convex ERM objective, which is often  optimized using first-order algorithms such as stochastic gradient descent, but can be plagued by \emph{poor initialization}, such that the algorithm may get stuck at a sub-optimal stationary point.
Although logistic functions are popular in the context of other machine learning problems such as classification, this paper presents the first consideration of adapting the logistic policies for offline policy learning problems. Details on the convex and logistic relaxation approaches can be found in Appendix \ref{app:relaxation}.

A main contribution of this paper is using Bayesian optimization to solve the ERM problem \eqref{eq:ERM}. Specifically, we adapt the Gradient-based Adaptive Stochastic Search (GASS) algorithm
proposed in \cite{zhou2014gradient} tailored to non-convex, non-smooth objectives. 
On a high level, instead of optimizing the ERM problem from a single starting point in the policy class, as many algorithms such as  logistic relaxation do, this Bayesian approach optimizes a \emph{distribution} over the policy class and thus avoids local minima and achieves global robustness.
We will demonstrate the effectiveness of Bayesian optimization at solving the ERM problem on simulated and real datasets. In particular, policies learned via Bayesian optimization consistently outperform policies learned from the aforementioned convex and logistic relaxation approaches, and are impressively robust to noise.

\subsection{A Bayesian Optimization Approach}
This section formulates the Bayesian optimization approach in the context of maximizing the ERM objective \eqref{eq:ERM}, as adapted from \cite{zhou2014gradient}. The key idea is to smooth the objective by estimating its expected value over the policy class with respect to a parameterized smooth probability density function, which is then updated iteratively  such that the distribution gradually concentrates on the set of global optima of the ERM problem. Algorithm \ref{alg:GASS} summarizes our implementation.

For completeness, we briefly summarize the Bayesian optimization framework proposed in \cite{zhou2014gradient}, which is a gradient-based stochastic search algorithm generalizing the Robbins-Monro algorithm \citep{robbins1951stochastic,chung1954stochastic}. One can apply
this framework  to any non-differentiable non-convex optimization problem, as long as the objective function is bounded below, 
and its value can be efficiently queried at any point of the domain. This is the case for our ERM problem over linear decision boundaries.


To be precise, suppose that our goal is to solve $\max_{z}H(z)$, with $H(z)$  lower bounded. Denote $f$ as  a smooth density function parameterized by $\omega$ over the domain of $H$. Define $L(\omega)\stackrel{\Delta}{=}\int H(z)f(z;\omega)dz$ as the expected value.
We then turn to solving the  problem $\max_\omega L(\omega)$, which is \emph{smooth} in $\omega$. We immediately have  $\max L(\omega) \leq\max H(\omega)\stackrel{\Delta}{=}H^{\ast}$; the equality holds 
 if and only if the parameterized  distribution
$f(z;\omega)$  is only supported on the set of
global optima of $H(z)$. The algorithm alternates between drawing
samples $\{z_{j}\}$ from $f(z;\omega)$ and using the samples to estimate $L(\omega)$ and update
parameter $\omega$ in a quasi-Newton step, with the guarantee that $f(z;\omega)$
becomes more and more concentrated on the set of global optima of
$H$ under general conditions. 

Operationally, consider $f$ as a distribution in an exponential family,
i.e. $f(z;\omega) =\exp(\omega^{T}T(z)-\phi(\omega))$. In practice, $f(z;\omega)$ can be taken to be a normal distribution, for which we may view smoothing as a convolution. By direct algebra, we obtain the gradient and Hessian of $L(w)$ as follows,
\begin{equation}
\begin{aligned}
\nabla_{\omega}L(\omega) & =\mathbb{E}_{\omega}\left[H(z)T(z)\right]-\mathbb{E}_{\omega}\left[H(z)\right]\cdot\mathbb{E}_{\omega}\left[T(z)\right]\\
\nabla_{\omega}^{2}L(\omega) & =\mathbb{E}_{\omega}\left[H(z)(T(z)-\mathbb{E}_{\omega}\left[T(z)\right])(T(z)-\mathbb{E}_{\omega}\left[T(z)\right])^{T}\right]\\
 & -\mathbb{V}_{\omega}\left[T(z)\right]\cdot\mathbb{E}_{\omega}[H(z)],
\end{aligned}
\end{equation}
where $\mathbb{V}_{\omega}[T(z)] = \mathbb{E}[T(z)T(z)^T] - \mathbb{E}[T(z)]\mathbb{E}[T(z)]^T$.
 To maximize the objective $L(\omega)$, a natural idea is to apply the Newton update and successively approximate the root to its gradient $\nabla_{\omega}L(\omega)$ as follows:
 \begin{equation}
 \label{eq:newton}
     \omega \leftarrow \omega - (\nabla_{\omega}^{2}L(\omega))^{-1}(\nabla_{\omega}L(\omega)).
 \end{equation}
 
However, to ensure that the update direction in \eqref{eq:newton} 
is an \emph{ascent} direction of $L(w)$ (recall that our goal is to maximize $L(w)$), we need to have the adaptive step size $(\nabla_{\omega}^{2}L(\omega))^{-1}$ be negative semidefinite, which unfortunately may not always hold.  A common practice is to replace $\nabla_{\omega}^{2}L(\omega)$ by its second term $-\mathbb{V}_{\omega}\left[T(z)\right]\mathbb{E}_{\omega}[H(z)]$, and thus we have,
 \begin{equation}
 \label{eq:newton-2}
     \omega \leftarrow \omega + \frac{\big(\mathbb{V}_{\omega}[T(z)]\big)^{-1}}{\mathbb{E}_{\omega}[H(z)]}\nabla_{\omega}L(\omega).
 \end{equation}
We claim that $\mathbb{V}_{\omega}[T(z)]$ is positive semi-definite. In fact, $\mathbb{V}_{\omega}\left[T(z)\right]$ is the Fisher information matrix and can be rewritten as  $\mathbb{V}_{\omega}\left[T(z)\right]=\mathbb{E}_{\omega}\left[\nabla_{\omega}\ln f(z;\omega)(\nabla_{\omega}\ln f(z;\omega))^{T}\right]=\mathbb{E}_\omega\left[-\nabla_{\omega}^{2}\ln f(z;\omega)\right]$
\citep{rao1945information}.
For computational stability, we follow \cite{zhou2014gradient} and add a positive perturbation term when calculating the inverse, i.e. replacing $(\mathbb{V}_{\omega}[T(z)])^{-1}$  in \eqref{eq:newton-2} with  $\big(\mathbb{V}_{\omega}[T(z)]+\epsilon I\big)^{-1}$ for some small $\epsilon>0$.

However, one more problem remains with the update rule \eqref{eq:newton-2}, since $\mathbb{E}_\omega[H(z)]$ is not necessarily
non-negative. To address this, \cite{zhou2014gradient} propose to apply a $\omega$-dependent filtering function $S_\omega(\cdot)$ to $H(z)$, such that (i) $S_\omega(H(z))$ is positive for all $z$; and (ii) only $z$'s with large $H(z)$ values  are sampled with high probability. Specifically, define
\begin{equation*}
    S_{\omega}(H(z)) \stackrel{\Delta}{=} (H(z)-H_{lb})\frac{1}{1+\exp(-S_{0}(H(z)-\gamma_{\omega}))},
\end{equation*}
where $S_{0}$ is a large positive constant, $H_{lb}$ is a lower bound of $H(z)$, and $\gamma_{\omega}$
is the $(1-\rho)$-quantile of $H(z)$ with density $f(z;\omega)$ and a small number $\rho>0$. To estimate $\gamma_{\omega}$ in practice, we sample $\{z_{m}\}_{m=1}^M$ from $ f(z;w)$ and use the empirical quantile calculated from $\{H(z_{m})\}_{m=1}^M$.
In fact, $\frac{1}{1+\exp(-S_{0}(H(z)-\gamma_{\omega}))}$ performs as a  smooth approximation of the indicator function $\mathbf{1}\{H(z)\geq \gamma_w\}$. Thus $S_{\omega}(H(z))$ is positive and is noticeably large
only if $H(z)\geq \gamma_{\omega}$. Moreover, we note that the filtering function $S_{\omega}(y)$ is increasing in $y$ and continuous in $\omega$, such that the set of
optimal solutions satisfies $\arg\max_{z}S_{\omega'}(H(z)) \subset\arg\max_{z}H(z)$
 for any $\omega'$. 
Therefore, instead of optimizing $L(\omega)=\int H(z)f(z;\omega)dz$, we solve the following optimization problem:
\begin{equation}
    \max_\omega L^S(\omega;\omega') = \int S_{\omega'}(H(z))f(z;\omega)dz.
\end{equation}
Similar to before, we have the gradient and the Hessian of $L^S(\omega;\omega')$ as follows:
\begin{equation}
\begin{aligned}
\nabla_{\omega}L^S(\omega;\omega') & =\mathbb{E}_{\omega}\left[S_{\omega'}(H(z))T(z)\right]-\mathbb{E}_{\omega}\left[S_{\omega'}(H(z))\right]\cdot\mathbb{E}_{\omega}\left[T(z)\right]\\
\nabla_{\omega}^{2}L^S(\omega;\omega') & =\mathbb{E}_{\omega}\left[S_{\omega'}(H(z))(T(z)-\mathbb{E}_{\omega}\left[T(z)\right])(T(z)-\mathbb{E}_{\omega}\left[T(z)\right])^{T}\right]\\
 & -\mathbb{V}_{\omega}\left[T(z)\right]\cdot \mathbb{E}_{\omega}[S_{\omega'}(H(z))],
\end{aligned}
\end{equation}and following the quasi-Newton update rule discussed above, we choose $\omega'=\omega$ and update the parameter $\omega$ via
  \begin{equation}
 \label{eq:newton-3}
 \begin{split}
\omega  \leftarrow & \omega + \frac{\big(\mathbb{V}_{\omega}[T(z)]+\epsilon I\big)^{-1}}{\mathbb{E}_{\omega}[S_{\omega}(H(z))]}\nabla_{\omega}L^S(\omega; \omega)\\
& =  \omega + \big(\mathbb{V}_{\omega}[T(z)]+\epsilon I\big)^{-1}\bigg(
\frac{\mathbb{E}_{\omega}[S_{\omega}(H(z))T(z)]}{\mathbb{E}_{\omega}[S_{\omega}(H(z))]}
-\mathbb{E}_{\omega}[T(z)]
\bigg).
 \end{split}
 \end{equation}
In practice, at step $k$, to estimate the update direction in \eqref{eq:newton-3}, we draw $M$ samples $\{z_m\}_{m=1}^M$ from the distribution $f(z;\omega^{(k)})$ and use empirical means and variance to estimate terms in \eqref{eq:newton-3}. In addition, we apply  a step size sequence $\{\alpha_k\}$ to the update direction, which satisfies $\sum_{k=0}^{+\infty}\alpha_k=\infty$ to meet the algorithm convergence condition~\citep{zhou2014gradient}.

Algorithm \ref{alg:GASS} details our implementation of the algorithm adapted to solve the ERM problem in the policy class of linear decision boundaries. 
In particular, we search in the domain of $\Theta$ and have $H(\Theta)=\frac{1}{n}\sum_{i=1}^n \widehat{\Gamma}_i(\pi(X_i;\Theta))$; we
consider a Gaussian prior $\mathcal{N}(\mu,\{\sigma_{j}^{2}\}_{j=1}^{d})$ for the density function $f(\Theta;\omega)$, where $\omega =\begin{bmatrix}\frac{\mu_1}{\sigma_{1}^{2}}, \cdots, \frac{\mu_d}{\sigma_{d}^{2}},
-\frac{1}{2\sigma_{1}^{2}},
\cdots,
-\frac{1}{2\sigma_{d}^{2}}
\end{bmatrix}$. 
We also note that the exponentiation calculation in $S_{\omega}(H(\Theta)) =(H(\Theta)-H_{lb})\frac{1}{1+\exp(-S_{0}(H(x)-\gamma_{\omega}))}$ requires the stable version to avoid numerical overflow problems, for which we  use
$\frac{1}{1+\exp(a)}=\frac{\exp(-a)}{1+\exp(-a)}$ when $a$ is positive,
due to the fact that $S_{0}$ is a very large constant. 

\begin{algorithm}

\caption{Bayesian Optimization for ERM of Learning Linear Decision Boundaries, Adapted from \cite{zhou2014gradient}
}

\label{alg:GASS}

\begin{algorithmic}[1] 
\STATEx \textbf{Input:} Observational data $\{(X_i, A_i, Y_i)\}_{i=1}^n$.
\STATEx \textbf{Output:} $\Theta\in\mathbb{R}^{pd}$, which determines a policy with linear decision boundaries $\pi(x;\Theta)\in \arg\max_{a\in\mathcal{A}}\langle\theta_a, x\rangle$.
\STATE \textbf{Construct  $H(\Theta)$.}
\begin{enumerate}
    \item Use cross-fitting to obtain $\hat{\mu}(x,a)$ and $\hat{e}(x,a)$ for the expected values $\mathbb{E}[Y(a)|X=x]$ and propensities $\mathbb{P}[A=a|X=x]$.
    \item Construct doubly robust estimators $\widehat{\Gamma}_i(a)=\frac{Y_{i}-\hat{\mu}(X_{i}, a)}{\hat{e}(X_{i}, a)}\cdot \mathbf{1}\{A_{i}=a\}+\hat{\mu}(X_{i},a)$.
    \item For any policy with linear decision boundaries specified by parameter $\Theta$, we have $H(\Theta)=\frac{1}{n}\sum_{i=1}^n\widehat{\Gamma}_i(\pi(X_i;\Theta))$.
\end{enumerate}
\STATE \textbf{Initialization and Hyperparameter Setting for Bayesian optimization:}  $\epsilon=10^{-5}$, $\rho=0.05$, $\sigma^2=\left[100\right]^{p}\in \mathbb{R}^{p}$, $\sigma_{\min}^{2}=10^{-6}$, $\mu=($uniform$[-1,1])^{p}$, $M=10^3$, $S_0=10^6$, $\alpha_0=1$, $a=0.05$, $k = 1$, set $H_{lb}=\mbox{lower bound of }H$.
\WHILE{not converged} 
\STATE{Sample $\{\Theta_m\}_{m=1}^{M}$ from $\mathcal{N}(\mu,\text{diag}(\sigma^{2}))$} 
\STATE{Evaluate $H_{\Theta}=\big(H(\Theta_{1}), \dots, H( \Theta_{M})\big)\in \mathbf{R}^M$} 
\STATE{Calculate $\gamma=(1-\rho)$th quantile of $\{H(\Theta_{m})\}_{m=1}^M$} 
\STATE{Calculate weights $\beta=(H_{\Theta}-H_{lb})\times\frac{1}{1+\exp(-S_{0}(H_{\Theta}-\gamma))} \in \mathbf{R}^M$ and normalize $\beta\rightarrow \frac{\beta}{\text{sum}(\beta)}$}
\STATE{Define  $T=\begin{bmatrix}\Theta_1, \dots, \Theta_m\\\Theta_1^{2},\dots, \Theta_m^2\end{bmatrix}\in\mathbb{R}^{2p\times M}$; Calculate  $\overline{T}_{\omega}= T\cdot \begin{bmatrix}\frac{1}{M}\\\vdots\\\frac{1}{M}\end{bmatrix}$ and $\overline{T}_{S}=T\cdot \beta$}
\STATE{Calculate $\overline{V}=\frac{1}{n-1}(T-\overline{T}_{\omega})\cdot(T-\overline{T}_{\omega})^{T}\in\mathbb{R}^{2p\times2p}$}
\STATE{Calculate step size $\alpha_k=\alpha_0/k^\alpha$} and update $\omega\leftarrow \omega +\alpha_k (\overline{V}+\epsilon I)^{-1}(\overline{T}_S - \overline{T}_\omega) $
\STATE{Set $\mu=-\frac{\omega[0:p]}{2\times\omega[p+1:2p]}$ and $\sigma^2=\max{(-\frac{1}{2\times\omega[p+1:2p]},\sigma_{\min}^{2})}$}
\STATE{$k=k+1$}
\ENDWHILE 
\end{algorithmic}
\end{algorithm}

\subsection{Synthetic Dataset with Linear Ground Truth Decision Boundaries}
\label{sec:linear}
We next evaluate the linear policy learning algorithms on synthetic
data with both ground-truth linear and nonlinear decision boundaries. Simulated data grant us access to the counterfactuals and therefore the true optimal policy, which enables us to calculate the value of the optimal policy and hence the regret of any policy. We leave the performance evaluation on a real dataset to Secion \ref{sec:applications}.

We start by describing the data generating process for linear ground truth decision boundaries. There are
three actions which we label as $\{0,1,2\}$. Each feature vector
$x=(x_0, x_1)\in\mathbb{R}^{2}$ is generated i.i.d. from uniform distribution
on $[-0.5,0.5]^2$. 
The feature space is partitioned into three regions with linear boundaries
in the 2-dimensional space: 
\begin{itemize}
    \item Region 0: $x_1 > 0,~0.8x_1> x_0 + 0.1 $;
    \item Region 1: $0.8x_1 < x_0 + 0.1,~ x_0 + x_1 + 0.1 > 0 $;
    \item Region 2: $x_1 < 0,~x_0 + x_1 + 0.1 < 0$.
\end{itemize}
The expected rewards for $x$ depend \emph{non-linearly} on $x$. Specifically, rewards  are decided purely by which region
$x$ is in and are designed so that the optimal action in region
$j$ is action $j$, as shown in Table \ref{tab:reward_linear}. The observed reward is perturbed by a Gaussion noise $\mathcal{N}(0,4)$. Similarly, the propensities (action selection probabilities) also depend non-linearly on which region $x$ is in and are given by Table \ref{tab:propensities_linear}. Recall that the propensities are specified by the data collection mechanism but may not be known to policy learning algorithms. In practice, one can estimate the propensites via cross-fitting techniques from the  observational data \citep{athey2017efficient}.

\begin{table}[]
\begin{subtable}[h]{0.48\textwidth}
    \centering
        \begin{tabular}{|l | l | l| l|}
        \hline
         & Action 0 & Action 1 & Action 2\\
       \hline
        Region 0 & 1.0 & 0.6 & 0.4\\
        \hline
        Region 1 & 0.5 & 1.0 & 0.3\\
        \hline
        Region 2 & 0.5 & 0.8 & 1.0\\
        \hline
       \end{tabular}
       \caption{Expected rewards for features in the partitioned regions.}
       \label{tab:reward_linear}
\end{subtable}
    \hfill
\begin{subtable}[h]{0.48\textwidth}
        \centering
        \begin{tabular}{|l | l | l|l|}
        \hline
         & Action 0 & Action 1 & Action 2\\
        \hline
        Region 0 & 0.2 & 0.6 & 0.2\\
        \hline
        Region 1 & 0.2 & 0.6 & 0.2\\
        \hline
        Region 2 & 0.4 & 0.2 & 0.4\\
        \hline
       \end{tabular}
       \caption{Propensities for features in the partitioned regions.}
       \label{tab:propensities_linear}
\end{subtable}
\caption{Data generating process for linear ground truth decision boundaries.}
     \label{tab:dgp_linear}
\end{table}

To summarize, for each observation $i$, a feature vector $X_i$ is  uniformly sampled from  $[-0.5,0.5]^2$; depending on which region $X_i$ belongs to, an action $A_i$ is chosen with probability proportional to the propensities specified by Table \ref{tab:propensities_linear}; an observed reward $Y_i$ is generated according to  $\mathcal{N}(\mu(X_i, A_i), 4)$, where the expected reward $\mu$ is given by Table \ref{tab:reward_linear} for each region. The dataset consists of  $10000$ such i.i.d. tuples $(X_i,A_i,Y_i)$. 

\begin{figure}[]
    \centering
    \includegraphics[width=\textwidth]{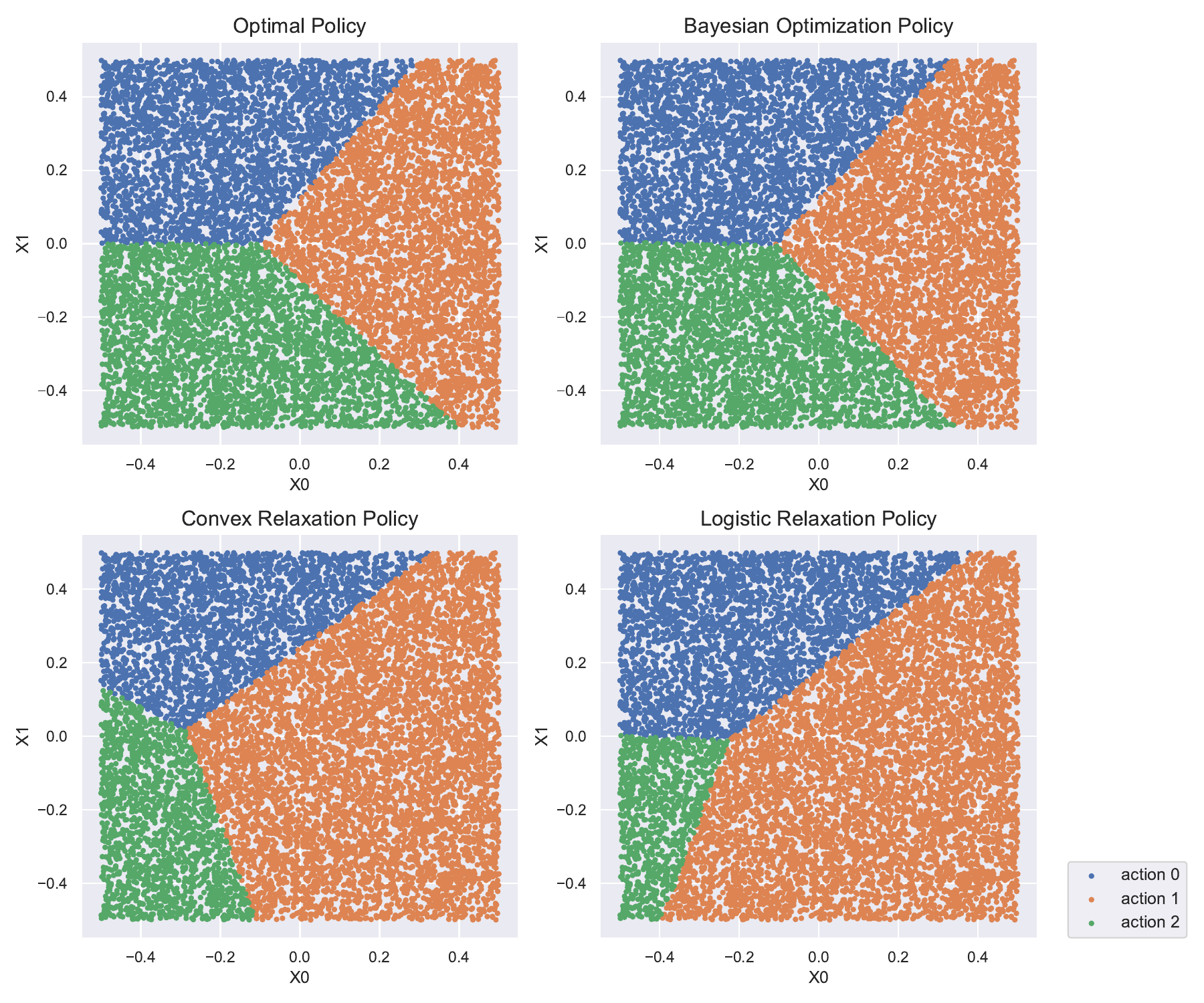}
    \caption{Decision boundaries learned from $10000$ samples with ground truth linear optimal decision boundaries.}
    \label{fig:linear}
\end{figure}

Figure \ref{fig:linear} presents the decision boundaries of the optimal policy and of learned policies from Bayesian optimization, convex relaxation, and logistic relaxation. One can see that the decision boundaries from Bayesian optimization closely resemble the optimal ones. The convex relaxation solves a problem that is a lower bound of the original objective, so its learned decision boundaries do not achieve exact recovery. Decision boundaries from the logistic relaxation also perform worse than those from Bayesian optimization, since as discussed, optimization through logistic relaxation may be plagued by bad initialization for non-convex problems. Table \ref{tab:results} further shows the aggregated results across $100$ simulation replicates, which demonstrates the effectiveness of Bayesian optimization over the two relaxation approaches at finding policies with minimal regret. 

\subsection{Synthetic Dataset with Nonlinear Ground Truth Decision Boundaries}
\label{sec:nonlinear}
Data with nonlinear ground truth decision boundaries is generated similarly as the linear case in Section \ref{sec:linear}. We  consider three actions $\{0,1,2\}$. Each feature vector $x=(x_0, x_1)\in \mathbb{R}^2$ is i.i.d. sampled from $\mbox{Uniform}[0,1]^2$. We  divide the feature space into three regions with nonlinear boundaries: 
\begin{itemize}
    \item Region 0: $x_0 <0.6,~x_1 >0.35 $;
    \item Region 1: $ ((x_0-1)/0.4)^2 + ((x_1-1)/0.35)^2 > 1,~x_0>0.6,~x_1 < 0.35$;
    \item Region 2:  $((x_0-1)/0.4)^2 + ((x_1-1)/0.35)^2 < 1$.
\end{itemize}
The region-dependent expected rewards and propensities are specified in Table \ref{tab:dgp_nonlinear}. Reward is realized by adding an i.i.d. Gaussian noise  $\mathcal{N}(0,4)$.

\begin{table}[]
\begin{subtable}[h]{0.48\textwidth}
    \centering
        \begin{tabular}{|l | l | l| l|}
        \hline
         & Action 0 & Action 1 & Action 2\\
       \hline
        Region 0 & 3.0 & 2.0 & 1.0\\
        \hline
        Region 1 & 1.5 & 2.5 & 1.5\\
        \hline
        Region 2 & -1.5 & 0 & 1.5\\
        \hline
       \end{tabular}
       \caption{Expected rewards for features in the partitioned regions.}
       \label{tab:reward_nonlinear}
\end{subtable}
    \hfill
\begin{subtable}[h]{0.48\textwidth}
        \centering
        \begin{tabular}{|l | l | l|l|}
        \hline
         & Action 0 & Action 1 & Action 2\\
        \hline
        Region 0 & 0.2 & 0.6 & 0.2\\
        \hline
        Region 1 & 0.2 & 0.6 & 0.2\\
        \hline
        Region 2 & 0.4 & 0.2 & 0.4\\
        \hline
       \end{tabular}
       \caption{Propensities for features in the partitioned regions.}
       \label{tab:propensities_nonlinear}
\end{subtable}
\caption{Data generating process for nonlinear ground truth decision boundaries.}
     \label{tab:dgp_nonlinear}
\end{table}
\begin{figure}
    \centering
    \includegraphics[width=\textwidth]{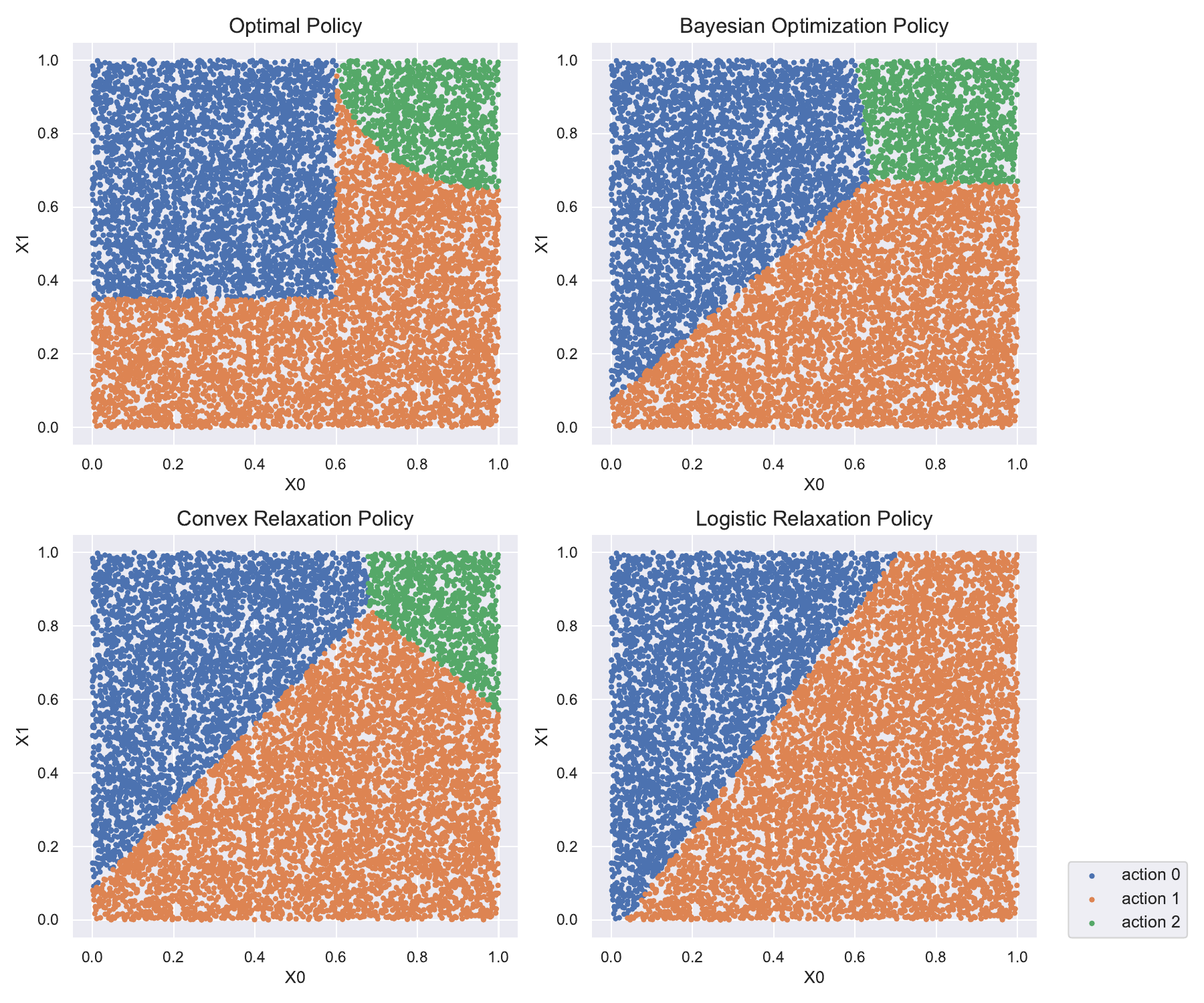}
    \caption{Decision boundaries learned from $10000$ samples with ground truth nonlinear optimal decision boundaries. Bayesian optimization yields the best linear approximation to the nonlinear decision boundaries.}
    \label{fig:nonlinear}
\end{figure}

\begin{table}[]
    \centering
    \begin{tabular}{|c|c|c|}
    \toprule
         &  Linear Ground Truth &  Non-Linear Ground Truth \\
         \midrule
        Optimal Policy &  1.0 &  2.5868\\
        Bayesian Optimization &  0.9881(0.0136)& 2.4588(0.0437)\\
        Convex Relaxation & 0.9561(0.0195)& 2.3310(0.0237)\\
        Logistic Relaxation & 0.9598(0.0404) & 2.2681(0.0372)\\
        \bottomrule
    \end{tabular}
    \caption{Values of learned policies. Results are aggregated over $100$ replicates. Numbers in the brackets denote standard deviations.}
    \label{tab:results}
\end{table}

We compare the decision boundaries of the optimal policy and policies with linear decision boundaries learned via Bayesian optimization, convex relaxation, and logistic relaxation in Figure \ref{fig:nonlinear}. By design, the optimal decision boundaries are nonlinear, which can not be exactly recovered by any policy in the class of linear decision boundaries. However, we see that Bayesian optimization presents a sensible linear approximation solution, and also achieves higher policy value compared to the two relaxation counterparts as shown in Table \ref{tab:results}. Convex relaxation also provides reasonable linear decision boundaries but performs worse empirically than the Bayesain optimization since it solves a lower bound of the original problem. Importantly, logistic relaxation performs worst in terms of both decision boundaries and policy values, since it easily gets stuck in a local optima without a global optimization perspective like the Bayesian approach. This example further demonstrates the Bayesian approach is fundamentally different from the logistic relaxation approach.

\section{Case Study: Personalizing Discount Offers on JD.com}
\label{sec:applications}
We now move on to evaluating our algorithms on a real dataset from JD.com about customer purchase behavior under coupon and discount incentives \citep{shen2020jd}.  

\subsection{Revenue Optimization with Targeted Discount Offers in Online Marketplaces}

On an online shopping platform such as JD.com, customers are constantly browsing products, a.k.a. stock keeping units or SKUs in short, which  they may be interested in purchasing. SKUs are either sold by third-party vendors or platform-owned brands, which we collectively refer to as sellers. Sellers have the option of offering discounts and coupons to customers, who make their decisions whether to purchase a particular item they are browsing based on discounts available to them as well as SKU characteristics. The net sales revenue, a.k.a. gross merchandise value (GMV) for a seller is the total sales value of the SKUs \emph{subtracted} by the discounts offered by the seller. Here a tradeoff arises for the seller: when discount is high, customers are more likely to make a purchase, but net revenue is smaller; when discount is low, fewer customers find it optimal to make a purchase, yielding more unrealized purchases which have zero value, but among realized purchases, net revenue is higher as the seller gave up less revenue through discounts. 

Therefore, a natural and important revenue management problem for sellers is to offer an optimal discount amount to customers in order to maximize the GMV. Importantly, because of heterogeneities in customer preferences and SKU characteristics, the optimal discount amount likely varies for different customer-SKU pairs. To the extent possible, sellers can offer \emph{personalized} discounts to improve GMV.

Learning such a decision rule from historical observational data poses several challenges. Crucially, we do \emph{not} know how particular discounts were offered or whether they were targeted, and at least some discounts were actively claimed by users, so the dataset suffers from the confounding issue that would bias conventional estimators of expected rewards. Fortunately, with detailed transaction-level data made available by JD.com \citep{shen2020jd}, we are able to apply our offline policy learning algorithms to this dataset to learn an optimal discount offering rule, with the goal of maximizing the GMV. We demonstrate that our interpretable policy learning framework solves this type of problems effectively, providing sellers with an interpretable and quantitative decision-making tool to offer personalized discounts to customers, improving significantly upon the GMVs achieved by both the baseline and naive estimators of rewards. 

\subsection{Learning Interpretable Personalized Discounts from JD.com Purchase Data}
\begin{figure}[!htb]
\minipage{0.3\textwidth}
  \includegraphics[width=\linewidth]{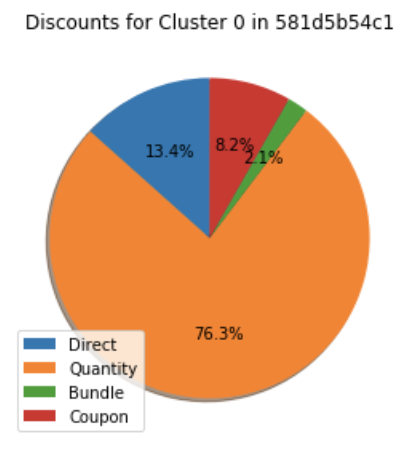}
 
\endminipage\hfill
\minipage{0.3\textwidth}
  \includegraphics[width=\linewidth]{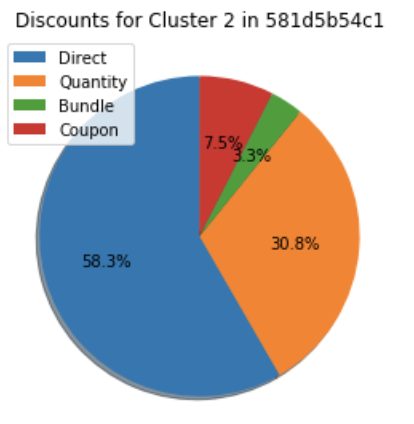}
  
\endminipage\hfill
\minipage{0.3\textwidth}%
  \includegraphics[width=\linewidth]{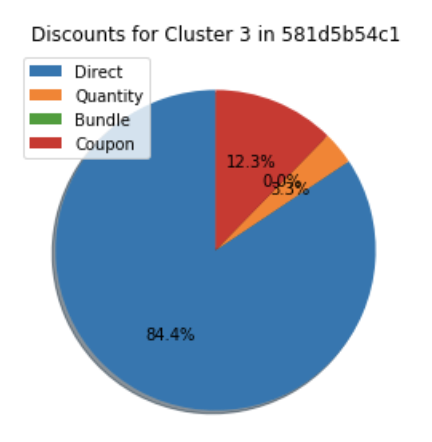}
\endminipage
\caption{Proportion of discounts present in purchases of SKU 581d5b54c1 by 3 different customer groups out of 10 clusters based on k-means. The heterogeneities suggest selection bias in discount offers.}\label{fig:pie chart}
\end{figure}

The JD.com dataset includes transaction-level purchase data, with information on the customer and SKU, quantity of purchase, as well as discounts applied at time of purchase. There are four possible types of discounts: direct discount (which are the most common), quantity discount (buy at least $n>1$ items to receive discount),  bundle discount (buy a specific set of SKUs to receive discount), and coupon discount (claimed by customers before purchase). Figure \ref{fig:pie chart} displays the proportion of each discount present in purchases of the same SKU by $10$ different groups of customers, which are obtained via $k$-means clustering based on customer and contextual features. Importantly, the proportion of available discounts varies considerably across these customer groups, and each group tends to receive a certain type of discount much more often than other types of discounts. This suggests that there is indeed \emph{selection bias} present in the assignment of discount offers. In fact, the data description in  \cite{shen2020jd} states that some customers already received personalized discounts based on past activities. This implies that a simple supervised learning approach using the observational dataset will result in biased estimates.

\begin{figure}
    \centering
    \includegraphics[width=.5\textwidth]{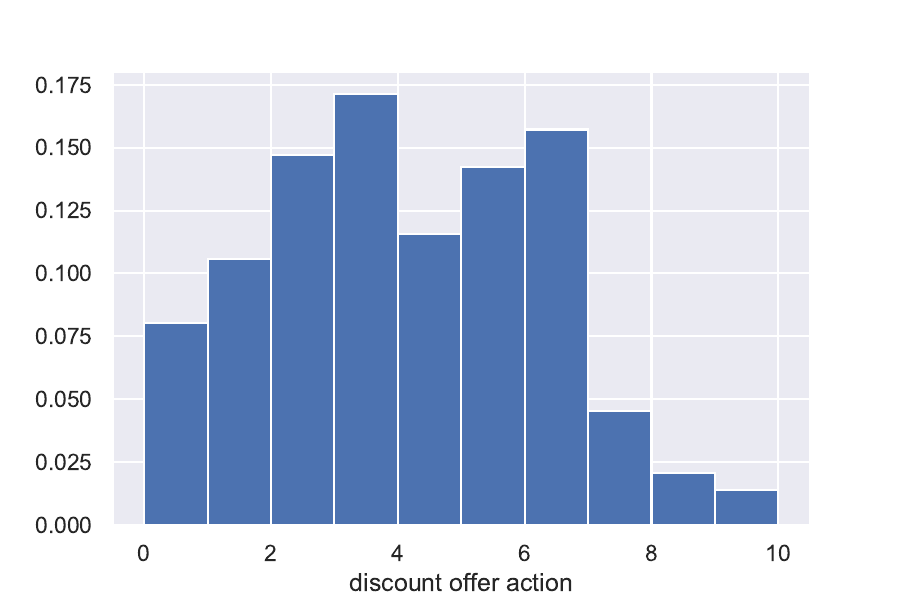}
    \caption{Distribution of actions (discount percentages) in the JD.com dataset.}
    \label{fig:jd_action}
\end{figure}

To concretely map the discount personalization problem onto our policy learning framework, we treat each customer-SKU pair as one sample. Contexts consist of customer characteristics such as gender, age, city tier, anonymized SKU features and price, as well as other information such as time of day. Actions are discount offers, and since we are interested in how the \emph{total} amount of discount offered to a customer for a particular SKU affects their purchase decision, the exact breakdown of the discount is less important. We thus combine the four discounts and discretize the total discount value into increments of $10$ percent of the original unit price, as our framework requires a \emph{discrete} action space. Figure \ref{fig:jd_action} shows the  distribution of actions in the data set. Finally, reward is the total net sales revenue for that SKU, i.e., price after applying discounts, multiplied by quantity. 

As the purchase data only contains observations where customers decided to make a purchase, we also need observations where customers decided against making a purchase, yielding zero revenue for the seller. For this we use the ``click dataset'', which records customers' browsing history of SKUs. By comparing with the purchase data, we can identify customer-SKU pairs in the click dataset that did not eventually lead to a purchase and subsample from these null observations. Unfortunately, the click dataset does not contain information on the discounts available to customers when they were viewing the products. We use the following strategy to infer each of the four discounts available for each sampled customer-SKU pair from the click dataset using observations from the purchase dataset, for which discount information is available. First, for each SKU, if a particular discount type only has one value (usually 0) in all observations in the purchase dataset, then we assign the value to that discount type for all sampled observations in the click dataset with that SKU. Otherwise, for each discount type for a particular SKU, we train a collection of parametric and non-parametric predictors, including linear regression, random forest, and matching, that output the discount amount per unit given customer and other contextual features, using observations from the purchase data containing that SKU. We then select the predictor with the smallest validation RMSE and use it to predict the amount of discounts available to customers browsing that SKU. See Table \ref{table:predicted discounts} for a few samples from the click dataset and predicted discounts output by the best predictor. Predicted discounts are then merged and discretized in the same way as observed discounts from the purchase dataset. The resulting set of observations that we use to learn a policy with linear decision boundaries consists of $\sim$100k customer-SKU pairs with features (contexts), discount level (action), and the net sales revenue (reward). See \cite{pauphilet2022robust} for a more robust procedure to impute discounts available to customers who did not make a purchase. \cite{pauphilet2022robust} also discusses the importance of incorporating the imputation procedure for \emph{partially observed} treatment assignments into the statistical estimation, although in this paper we focus on the optimization aspect of the problem.

\begin{table}[]
    \centering
    \begin{tabular}{|l|l|l|l|l|}
    \hline
sku\_ID                  &     a1b0f57464 &     2523d051fd &     66ab6d8963 &     dbd945bd02 \\
user\_ID                 &     6aaf2a5c3a &     02194ae5bc &     207a5b513a &     62416a0df3 \\
\hline
direct\_discount         &           8.11 &             80 &        3.42153 &              0 \\
rmse\_direct\_discount    &      0.0672924 &      0.0422325 &       0.147431 &            0 \\
best\_predictor\_direct   &  random\_forest &          matching &         linear &   single value \\
\hline
quantity\_discount       &              0 &           89.4 &        6.17357 &           3.78 \\
rmse\_quantity\_discount  &            0 &      0.0977727 &       0.278181 &       0.156358 \\
best\_predictor\_quantity &   single value &          matching &         linear &  random\_forest \\
\hline
bundle\_discount         &              0 &            104 &           23.5 &              0 \\
rmse\_bundle\_discount    &            0 &       0.264352 &       0.771073 &            0 \\
best\_predictor\_bundle   &   single value &          matching &  random\_forest &   single value \\
\hline
coupon\_discount         &       0.496766 &           25.2 &              0 &        0.21534 \\
rmse\_coupon\_discount    &       0.152163 &       0.504334 &      0.0116617 &       0.131212 \\
best\_predictor\_coupon   &         linear &  random\_forest &       logistic &         linear \\
\hline
\end{tabular}

    \caption{Imputed discounts for four customer-sku pairs in the click dataset. For each discount type, we report the best predictor and associated RMSE.}
    \label{table:predicted discounts}
\end{table}

\begin{figure}
    \centering
    \includegraphics[width=1\textwidth]{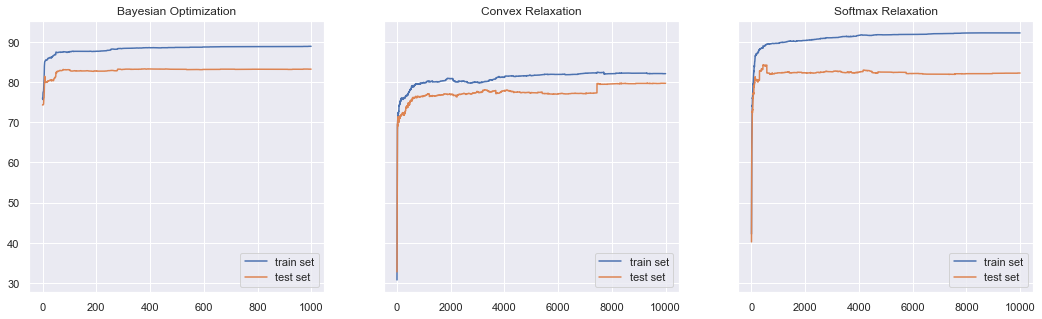}
    \caption{Average train and test policy values of three optimization approaches during training. Final test policy values: Platform Baseline: 44.20; Bayesian: 83.20;
Convex Relaxation: 79.69;
Softmax Relaxation: 82.23.}
    \label{fig:jd_policies}
\end{figure}

With this observational dataset, we apply our offline policy learning algorithms to learn linear decision boundaries that choose the optimal discount level to offer to a particular customer-SKU pair. 
In order to evaluate  a learned policy, ideally we need the actual purchase decision of a sampled customer when they are presented with a particular SKU and the discount level suggested by the policy, so that we can calculate the net sales revenue. Unfortunately, these counterfactuals are not available in our observational dataset. To overcome this problem, 
we split the data into training set and test set with ratio $2:1$. 
For each customer-SKU pair in the test set with a particular discount level suggested by our policy learned on the training data, we predict the counterfactual customer response and estimate the net revenue using doubly robust reward estimators that are cross-fitted on the test set \citep{zhou2018offline}.
We compare the total net sales revenue generated by policies learned from Bayesian optimization, convex relaxation, and logistic relaxation, with the baseline discount offer policy currently used by the platform. Figure \ref{fig:jd_policies} shows the train and test curves of policy values for the three optimization approaches. The final numbers suggest that personalized discount offers based on the policy learned using Bayesian optimization outperforms the baseline policy used by the platform by 88.2\% in terms of average net sales revenue (GMV), followed by the policy from logistic relaxation with 86.0\% and then by the policy from convex relaxation with 80.3\%. While in the most ideal situation, we would be able to validate and assess our algorithms in a randomized experiment, we find these results based on counterfactual estimations highly encouraging.

As one of the main advantages of our approach is the interpretability of learned policies, we discuss some findings regarding what practical and simple recommendations we can make to sellers to help them improve GMV. For example, the feature ``Attribute 2'' for an SKU is a quality indicator (higher value means better quality), and our learned policy suggests that when this feature is large, the optimal discount level generally should be greater than 0\% but less than 10\%, \emph{especially} if the user belongs to age groups of 16-25 years or $>56$ years, or if the user is a ``plus'' member. This is plausible, since higher quality goods have higher utility and hence require less discounts to nudge customers to make a purchase decision in general. Moreover, since plus members enjoy free shipping and unconditional returns, their price elasticity of demand is smaller than others; similarly, customers in certain age groups tend to place greater importance on quality vs. price or may have particular brand loyalties, and are thus less price-sensitive for such products.  

\section{Concluding Remarks}
In this paper, we apply the framework of offline policy learning to the important problem of interpretable personalization. We focus on the class of policies with linear decision boundaries, which allows sophisticated (non-linear) estimation schemes for the reward functions. We propose using cross-fitted doubly-robust estimators from the causal inference literature for reward estimation, and a Bayesian optimization scheme to accurately solve the associated non-convex, non-smooth ERM problem. The Bayesian approach is shown to outperform relaxation approaches in simulations and on real data. Moreover, 
through the application to a customer purchase dataset from JD.com, we demonstrate that linear decision boundary policies can provide insights and actionable recommendations to the platform by taking into account the impact of customer and product heterogeneities on sellers' revenues, helping to potentially improve the gross merchandise value (GMV) by as much as 88\%. We see this as a major advantage of our framework over black-box approaches, and we believe that it could be useful in a wide range of applications with personalized decision-making problems where interpretability is critical. 

There are at least two interesting directions we see for future works based on the current framework. The first is a more systematic integration of the discount imputation procedures into the optimization as well as statistical estimation workflows. As noted by \cite{pauphilet2022robust}, this corresponds to a \emph{partially} observable treatments setting with non missing-at-random (MAR) data, which is common in operations management problems. The resulting problem concerning statistical efficiency and regret minimization remains to be studied thoroughly. Second, as detailed in \cite{ettl2020data}, there are often time-dependent inventory constraints for sellers on online shopping platforms, so the problem of inventory management creates a potential trade-off with the problem of 
revenue management. Consequently, an optimal personalization policy for such online platforms should also take into account such factors as product availability, inventory health, and shipping time and cost. We consider this aspect to be a very interesting future direction for policy learning-based personalization in operations management.

\textbf{Acknowledgements} The authors would like to thank Ruohan Zhan and Ying Jin for helpful discussions and excellent research assistance.

\bibliographystyle{informs2014} 
\bibliography{reference.bib} 


\newpage

\begin{appendices}

\section{Relaxation Approaches to Solving the ERM Problem}

\label{app:relaxation}

\subsection{Convex Relaxation}
The convex relaxation approach seeks a differentiable, concave lower bound of the objective
function of the ERM problem, and maximizes this concave surrogate instead. Our formulation
is inspired by the convex relaxation technique commonly used in weighted
classification (\cite{shalev2014understanding}). 

For notation convenience, denote $\Psi(x,a^{j})=x\otimes a^{j}\in\mathbb{R}^{pd}$. Then for a policy $\pi$ specified by parameters of linear decision boundaries $\Theta$, we have $\pi(x;\Theta)=\arg\max_{j\in[d]}\langle\Theta,\Psi(x,a^{j})\rangle$.
By definition,
\begin{align}
\langle\Theta,\Psi(x,\pi(x;\Theta))\rangle\geq & \langle\Theta,\Psi(x,a')\rangle, \quad\forall a'.
\end{align}
 Considering observation $i$ with reward estimator $\hat{\Gamma}_i(a)$,
choose $a_i^{\ast}=\arg\max_{a\in\mathcal{A}}\hat{\Gamma}_i(a)$ as the best action with respect to $\hat{\Gamma}_i$. We  
have 
\begin{align}
\hat{\Gamma}_i(\pi(X_i;\Theta))  \geq\hat{\Gamma}_i(\pi(X_i;\Theta)) -\langle\Theta,\Psi(X_i,\pi(X_i;\Theta))-\Psi(X_i,a_i^{\ast})\rangle.
\end{align}
We can lower bound the right hand side of the above by
\begin{align}
\min_{a'\in\mathcal{A}} \hat{\Gamma}_i(a')-\langle\Theta,\Psi(X_i,a')-\Psi(X_i,a_i^{\ast})\rangle & :=V(\Theta,(X_i,\hat{\Gamma}_i)).
\end{align}

We see that the surrogate value function $V(\Theta,(X_i,\hat{\Gamma}_i))$ is associated
with covariate $X_i$ and reward $\hat{\Gamma}_i$ (which determines $a^{\ast}_i$) and 
 satisfies $V(\Theta,(X_i,\hat{\Gamma}_i))\leq \hat{\Gamma}_i(\pi(X_i;\Theta))$. Since by definition, $V$ is the minimum of linear functions in $\Theta$,
it is concave in $\Theta$ and its Lipschitz constant is $\min_{a'\in\mathcal{A}}\|\Psi(x,a')-\Psi(x,a)\|$.
Thus by solving the convex problem $\max_{\Theta}\frac{1}{n}\sum_{i}V(\Theta,(X_{i},\hat{\Gamma}_{i}))$
 we can obtain a lower bound on the optimal value of the ERM problem. To solve the surrogate problem, we can use stochastic subgradient descent. At each  step $k$, we sample $(X_{i},\hat{\Gamma})$ from the dataset,
and find 
\begin{align}
\hat{a}\in\arg\min_{a'\in\mathcal{A}} \hat{\Gamma}_{i}(a')-\langle\Theta,\Psi(X_{i},a')-\Psi(X_{i},a^{\ast}_i),\rangle
\end{align}
which is used for  updating  $\Theta^{(k+1)} =\Theta^{(k)}-\alpha_k(\Psi(X_{i},\hat{a})-\Psi(X_{i},a_{i}^{\ast}))$ with step size $\alpha_k$.

 \subsection{Logistic Policy Class Relaxation}

Instead of considering a deterministic policy class with linear decision boundaries, we can relax the original ERM problem to seaching solutions in the class of  \emph{logistic policies}. Given a feature vector $x\in\mathbb{R}^p$ and action parameters $\Theta\in\mathbb{R}^{pd}$, rather than selecting the action deterministically corresponding to the
largest inner product, as does by
$\pi(x;\Theta) \in\arg\max_{a\in\mathcal{A}}\langle\theta_{a},x\rangle
$, a logistic policy $h$ applies a softmax transformation to the inner products
and obtains a probabilistic policy:
\begin{align}
h(x;\Theta) & =\begin{bmatrix}\frac{\exp(\theta_{a^1}^{T}x)}{\sum_{j}\exp(\theta_{a^j}^{T}x)}\\
\vdots\\
\frac{\exp(\theta_{a^d}^{T}x)}{\sum_{j}\exp(\theta_{a^j}^{T}x)},
\end{bmatrix}
\end{align}
which specifies the action selection probabilities.
The solution of the ERM optimization problem over this different class
of policies does not yield a lower bound on the objective, but solving it is differentiable, which allows us  to directly apply techniques of 
stochastic gradient descent (SGD) to optimize it. Empirical performance of such logistic relaxation approach shows that its decision
boundaries are often close to those of linear policies. This is because the learned logistic policy returns  a probability vector  for each context $x$ that almost always has a dominant coordinate, so that with probability close to 1, the policy will take the corresponding action. An additional advantage of logistic policies is that they are amenable to nonlinear extensions, where the inner products are replaced with nonlinear functions such as neural networks. The resulting ERM problem can similarly be solved using SGD, and the resulting policy trades interpretability for the ability to learn nonlinear decision boundaries in the feature space. 

\end{appendices}

\end{document}